# Detection of REM Sleep Behaviour Disorder by Automated Polysomnography Analysis

Navin Cooray[1], Fernando Andreotti[1], Christine Lo[2], Mkael Symmonds[3], Michele T.M. Hu[2], & Maarten De Vos[1]

[1]University of Oxford, Institute of Biomedical Engineering, Dept. Engineering Sciences, Oxford, UK.

[2]Nuffield Department of Clinical Neurosciences, Oxford Parkinson's Disease Centre (OPDC), University of Oxford, UK.

[3]Department of Clinical Neurophysiology, Oxford University Hospitals, John Radcliffe Hospital, University of Oxford, UK.

Author to whom all correspondence should be directed is NC (navin.cooray@eng.ox.ac.uk)

## Abstract

Objective: Evidence suggests Rapid-Eye-Movement (REM) Sleep Behaviour Disorder (RBD) is an early predictor of Parkinson's disease. This study proposes a fully-automated framework for RBD detection consisting of automated sleep staging followed by RBD identification.

Methods: Analysis was assessed using a limited polysomnography montage from 53 participants with RBD and 53 age-matched healthy controls. Sleep stage classification was achieved using a Random Forest (RF) classifier and 156 features extracted from electroencephalogram (EEG), electrooculogram (EOG) and electromyogram (EMG) channels. For RBD detection, a RF classifier was trained combining established techniques to quantify muscle atonia with additional features that incorporate sleep architecture and the EMG fractal exponent.

Results: Automated multi-state sleep staging achieved a 0.62 Cohen's Kappa score. RBD detection accuracy improved by 10% to 96% (compared to individual established metrics) when using manually annotated sleep staging. Accuracy remained high (92%) when using automated sleep staging.

Conclusions: This study outperforms established metrics and demonstrates that incorporating sleep architecture and sleep stage transitions can benefit RBD detection. This study also achieved automated sleep staging with a level of accuracy comparable to manual annotation.

Significance: This study validates a tractable, fully-automated, and sensitive pipeline for RBD identification that could be translated to wearable take-home technology.

## Keywords

## Highlights

- RBD detection benefits from an ensemble of metrics that incorporate sleep architecture.
- RBD detection remains successful using both manual and automated sleep scoring.
- Cohort outnumbers similar studies with 53 RBD participants and 53 aged-matched healthy controls.

## Acknowledgments

This research was supported by the Research Council UK (RCUK) Digital Economy Programme (Oxford Centre for Doctoral Training in Healthcare Innovation -- grant EP/G036861/1), Sleep, Circadian Rhythms & Neuroscience Institute (SCNi -- 098461/Z/12/Z), Rotary Foundation, National Institute for Health Research (NIHR) Oxford Biomedical Research Centre (BRC) and the Engineering and Physical Sciences Research Council (EPSRC -- grant EP/N024966/1). The content of this article is solely the responsibility of the authors and does not necessarily represent the official views of the RCUK, SCNi, NIHR, BRC or the Rotary Foundation.

## Conflict of interest

None of the authors have potential conflicts of interest to be disclosed.

## 1. Introduction

Rapid-Eye-Movement (REM) Sleep Behaviour Disorder (RBD) is a parasomnia first described in 1986, characterised by loss of normal muscle atonia and dream-enactment motor activity during REM sleep (Schenck et al. 1986; Kryger et al. 2011). There is clear evidence that RBD is a precursor to Parkinson's Disease (PD), Lewy Body disease (LBD), and multiple system atrophy (MSA), preceding them by years, potentially decades (Schenck et al. 2013). Therefore, an accurate RBD diagnosis would provide invaluable early detection and insights into the development of these neurodegenerative disorders.

A definitive diagnosis of RBD, standardised by the current International Classification of Sleep Disorders (ICSD3) requires polysomnography (PSG) evidence of REM sleep without atonia (RSWA), and involves concurrent recording in sleep of electroencephalogram (EEG), electrooculogram (EOG), and electromyogram (EMG) signals.

There are numerous methods in the literature that utilise PSG recordings to classify REM stages either specifically (Kempfner et al. 2012; Imtiaz and Rodriguez-Villegas 2014; McCarty et al. 2014; Yetton et al. 2016) or as part of multi-sleep-stage classification (Virkkala et al. 2008; Güneş et al. 2010; Fraiwan et al. 2012; Liang et al. 2012; Bajaj and Pachori 2013; Kempfner et al. 2013b; Khalighi et al. 2013; Lajnef et al. 2015; Sousa et al. 2015). Many of these automated sleep scoring algorithms produce results that are comparable to expert annotation of PSG recordings from young healthy controls. However, very few of those validated automated scoring algorithms were designed for older individuals, let alone those who suffer from sleep disorders that can exhibit very different EEG characteristics (Iber et al. 2007; Luca et al. 2015). Nonetheless, manual sleep scoring remains the clinical gold-standard to date.

Traditional RBD diagnosis requires repeated episodes of RSWA, either visually identified with PSG or presumed to occur based on reports of dream-enacting behaviour (Sateia 2014). To provide clarity and consistency, Lapierre and Montplaisir (1992) first proposed scoring rules to quantify abnormal EMG tonic and phasic activity (Lapierre and Montplaisir 1992). This method was further developed (Dauvilliers et al. 2007; Montplaisir et al. 2010; Fulda et al. 2013), but still required manual visual inspection to distinguish tonic and phasic movement. Automated RBD detection through RSWA has been proposed in a number of papers (Burns et al. 2007; Kempfner et al. 2013a; Frauscher et al. 2014; Frandsen et al. 2015). Burns *et al.* (2007) used EMG variance to develop a metric called the Supra-Threshold REM EMG Activity Metric (STREAM). This metric was calculated by measuring the EMG variance during REM epochs and comparing it to a threshold calculated during non-REM (NREM) epochs. Kempfner *et al.* (2013) described a technique that requires three EMG channels and from each channel a single feature was calculated by comparing the mean envelope of a mini-epoch to the minimum envelope of the entire epoch. A one-class support vector machine was then used for detecting anomalous epochs, using manually and automatically annotated sleep staging (Kempfner et al. 2013a, 2014). The REM atonia index developed by Ferri *et al.* (2008) provides a score for the level of atonia by analysing the distribution of the filtered, rectified and averaged EMG amplitudes. However recordings may be imbued with noise and artefacts that can distort the calculation, for this purpose the corrected atonia index score (Ferri et al. 2010) was proposed. Frauscher *et al.* (2014) achieved automatic RBD detection by quantifying motor activity based on an index score measuring phasic and tonic activity from the mentalis and flexor digitorum superficialis muscle. Lastly, Frandsen *et al.* (2015) made use of a sliding window and a threshold to identify motor activity. This was then used to derive features that depict the number of motor activity events, quantified by duration and percentage of REM epochs, to distinguish RBD from other individuals. Despite the abundance of studies on RBD detection, these objective metrics have been applied to relatively small RBD cohorts to date (ranging from 10 to 31 patients), achieving variable sensitivity and specificity (0.74-1.00 and 0.71-1.00, respectively) (Burns et al. 2007; Ferri et al. 2008, 2010; Kempfner et al. 2013b; Frauscher et al. 2014; Kempfner et al. 2014; Frandsen et al. 2015). In a preliminary study we showed that using an ensemble of established techniques improved RBD detection, and can be further enhanced with the incorporation of sleep architectural features (Cooray et al. 2018).

In this study we propose a fully automated pipeline for RBD detection. The framework combines REM detection and abnormal EMG quantification in a cohort of age-matched healthy and RBD-diagnosed participants. Additionally, novel features that quantify the EMG fractal exponent ratio between sleep stages are combined with sleep architecture metrics to further improve RBD detection.

## 2. Data

PSG recordings from individuals diagnosed with RBD and age-matched healthy controls (HCs) were collated using several sources, detailed in Table 1. PSG recordings of 53 HC individuals were obtained from the Montreal Archive of Sleep Studies (MASS) cohort 1, subset 1 database [dataset] (O'Reilly et al. 2014). The combined RBD dataset consisted of 22 RBD participants from the Physionet Cyclic Alternating Pattern (CAP) sleep database [dataset] (Goldberger et al. 2000; Terzano et al. 2001) and 31 participants from a private database acquired by our local partners from the John Radcliffe (JR) hospital, Nuffield Department of Clinical Neurosciences at the University of Oxford. Participants from the JR dataset have been clinically diagnosed with idiopathic RBD with no concurrent PD, LBD, or MSA. While the CAP sleep database simply states that the participants are affected by RBD. Only two RBD participants from the JR dataset were taking Clonazepam to treat their condition and nine participants were taking antidepressants (Citaloprasm, Venlafaxine, and Sertraline) to treat movements during REM. The Apnoea Hypopnea Index (AHI) for all participants does not exceed moderate levels, unfortunately the AHI for the CAP database is unknown. The MASS dataset provided an AHI score for all participants, only three healthy controls had a score greater than 15, considered moderate. The AHI score of RBD participants was not measured if obstructive sleep apnoea (OSA) was previously excluded or oxygen saturation monitoring was unremarkable. The measured AHI scores of RBD participants were all less than 7.1 and were considered mild and unremarkable. Five RBD participants used a continuous positive airway pressure ventilator during their PSG, indicating they had OSA. This study complied with the requirements of the Department of Health Research Governance Framework for Health and Social Care 2005 and was approved by the Oxford University hospitals NHS Trust (HH/RA/PID 11957). The JR dataset comprised two nights of full PSG recordings for each participant, but for the purposes of this study only the second night was used (where available). The CAP database and the recordings from the JR hospital were combined to utilise the entire MASS dataset, while also evaluating how well this study generalises over numerous datasets annotated by different institutions. Once more by using openly available datasets (MASS and CAP), this study can be reproduced with the toolbox provided at https://github.com/navsnav/RBD-Sleep-Detection.

All PSG recordings were annotated by an expert who assigned a sleep stage for every epoch using either Rechtschaffen and Kales (R&K) or American Academy of Sleep Medicine (AASM) guidelines. Recordings annotated using R&K were converted to AASM, simply by assigning S3 and S4 to N3, while S0, S1 and S2 relabelled as W, N1 and N2, respectively. This study focused on three PSG signals in order to test the feasibility of developing an automated pipeline that could be translated into a take home-device with a limited number of channels:

- 1 EEG (either C4-A1, C3-A2 or C1-A1, listed in preferential order)
- 1 EOG (delta of ROC and LOC)
- 1 EMG (chin – submentalis)

Numerous studies have been able to emulate human performance in automated sleep staging using a limited number of channels, including one study that uses a single channel (Supratak et al. 2017; Andreotti et al. 2018; Chambon et al. 2018).

## 3. Method

### 3.1 Pre-processing

PSG recordings were first pre-processed to reduce noise and the effect of artefacts. To ensure consistency between the various recordings, all EEG, EOG and EMG signals were resampled at 200Hz. The EEG and EOG signal were pre-processed with a $500^{th}$ order band pass finite impulse response (FIR) filter with a cut-off frequency of 0.3Hz and 40Hz. The EMG signal was filtered with a $500^{th}$ order notch filter at 50Hz and 60Hz (because data is sourced from either Europe or Canada), in addition to a $500^{th}$ order band pass FIR filter between 10 and 100Hz.

### 3.2 Feature Extraction

The literature on automated sleep staging describes numerous features, which provided the basis for this study and are summarised in Table 2. For this study the pre-processed EEG, EOG, and EMG signals were segmented into 10 second mini-epochs in order to calculate features for each 30-second epoch, a technique often used for sleep stage classification literature (Güneş et al. 2010; Koley and Dey 2012; Liang et al. 2012; Lajnef et al. 2015; Yetton et al. 2016).

## 3.3 Automated Sleep Stage Classification

The well-known Random Forest (RF) algorithm (Breiman 2001) was trained for classifying 30-second epoch into one of the five classes described in the AASM norm (i.e. Wake, N1, N2, N3, REM). A RF consists of many decision trees and the classifier designs each node of a tree using a random subset of features, which provides resistance to over-fitting. A total of 156 features were derived using EEG, EOG, and EMG signals and for this study all features were used to train an RF classifier. In this study the number of trees was set at 500 and the number of randomly selected features for node branching was chosen as m_try=$\sqrt{M}$ (rounded down), where $M$ is the total number of features (in this study $M$=156 and m_try=12). The random forest was designed to provide a multi-stage sleep stage classifier using all features mentioned in section 3.2. The performance of the classifier was evaluated using macro-averaged sleep stage accuracy, sensitivity, and specificity by using 10-fold cross-validation with an even split between healthy and RBD participants. Multi-stage sleep classification was assessed by Cohen's Kappa (Cohen 1960), which provides a measure of the agreement between two-raters (in this case between a human and an automated algorithm).

## 3.4 RBD Detection

Once REM sleep has been identified, RBD diagnosis mandates the visual identification of REM sleep without atonia (RSWA) (Sateia 2014). Automated RBD detection was implemented based on established techniques that quantify RSWA (Burns et al. 2007; Ferri et al. 2010; Frandsen et al. 2015). The ability of these established metrics to automatically distinguish RBD individuals was evaluated using both automatically and manually annotated REM epochs. Additionally the correlation of these metrics derived using manually and automatically annotated sleep stages were measured to analyse the impact of automatic multi-stage sleep classification on RBD detection.

Additional RF classifiers were also used for RBD detection. One was trained with the three established RSWA metrics (with 500 trees, M = 4, and m_try = 2) using a 10-fold cross-validation scheme with both manually annotated and automatically classified sleep stages. In addition to the established RSWA metrics, we proposed the use of new relevant features, ones that incorporate sleep architecture (Massicotte-marquez et al. 2005) and the EMG fractal exponent (Krakovská and Mezeiová 2011). Another RF classifier (500 trees, M = 10, and m_try = 3) was trained and tested based on 10 features, namely the three established RSWA metrics, the mean fractal exponent relative to REM (during N3 and N2), percentage of N3 sleep (excluding wake), sleep efficiency and the atonia index relative to REM (during N3 and N2).

# 4. Results

The following results are presented in three sections, detailing 1) automated sleep stage classification, 2) correlation of EMG and sleep architecture metrics using automatically/manually annotated sleep stages, and 3) RBD detection.

## 4.1 Automated Sleep Stage Classification

Overall, the automated sleep stage classifier performed well, with an accuracy on the combined (healthy control and RBD) dataset achieving a multi-stage agreement score of 0.62, considered a substantial agreement (Landis and Koch 1977). When analysed individually, the HC cohort attained a substantial agreement score of 0.73, compared to the moderate score of 0.54 for the RBD cohort.

Robust detection of REM sleep stages is crucial for accurate diagnosis of RBD. Overall, the classifier achieved 0.93±0.05 accuracy, 0.64±0.31 sensitivity, and 0.97±0.04 specificity for classification of REM sleep stages on the combined dataset. There was a substantial difference in the REM detection sensitivity between the healthy and RBD cohort, at 0.83±0.18 and 0.45±0.30, respectively. However the specificity for REM detection remains very high for both cohorts, which will prove pivotal in avoiding false positive RSWA results. A summary of multi-stage sleep classification performance for each cohort are provided in Table 3 (HC) and Table 4 (RBD).

The results from multi-stage classification are detailed in the confusion matrices shown in Figure 1(a) and (b). Across all sleep stages there is greater rate of misclassification in participants with RBD compared to HCs. In total 22% of annotated N2 epochs for RBD individuals are misclassified, compared to 10% in HCs. Perhaps the greatest differentiator in performance for the RBD cohort, is the misclassified REM epochs, which are misclassified as N2 and W by 32% and 17% (substantially higher than the HC at 8.6% and 2.0%, respectively).

## 4.2 Metric Correlation for Automated and Annotated Sleep Staging

The impact of automated sleep staging on quantified EMG and sleep architecture metrics was evaluated by measuring the Pearson correlation (pairwise) between scores derived from manually annotated and automatically classified sleep stages, depicted in Figure 2 (a) to (i).

## 4.3 RBD Detection

The ability to discriminate individuals with RBD from healthy controls using individual metrics and two RF classifiers (trained on established metrics and established metrics supplemented with additional features) is depicted in Table 5. The best performance was attained by the RF classifier trained on a combination of established metrics supplemented by features that incorporate sleep architecture with an accuracy, sensitivity, and specificity of 0.96, 0.98 and 0.94, respectively. Additionally this table depicts the performance of RBD detection using automated sleep staging, where the performance is only marginally lower with manual sleep annotation (accuracy, sensitivity and specificity of 0.92, 0.91 and 0.93, respectively). From the classifier we can derive the feature importance using the permuted delta prediction error, where REM atonia index proves the most important as shown in Figure 3, followed by the N3 sleep ratio and the atonia index ratio (N3/REM).

## 5. Discussion

The goal of this study was to validate a fully-automated pipeline for identifying individuals with RBD. Firstly, this was achieved using automated multi-stage sleep classification, which had a high accuracy for detection of REM sleep stages compared to the gold standard of human expert classification (Figure 1). Despite a drop in performance for classification of REM stages in RBD individuals, specificity of REM stage detection remained high, and EMG quantification metrics based on automated sleep staging were shown to be highly correlated with manual annotations (Figure 2). As a result, automated RBD detection can be successfully achieved using automated sleep staging and established EMG metrics. Moreover, it was shown that RBD identification can be further enhanced using additional features that combine sleep architecture and EMG movement quantification (Table 5). This performance of RBD detection remained high using automated sleep staging, once again due to sleep staging performance. This was all achieved using a limited montage of a single EEG, EOG and EMG channel, enabling this automated system to be directly incorporated into lightweight wearable technology.

There is a significant degree of variability in manual sleep staging even with highly experienced scorers, with estimates of human inter-rater variability in sleep by Cohen's Kappa of 0.68 to 0.76 (Danker-Hopfe et al. 2004, 2009). This variability can be further compounded by sleep disorders, where agreement scores can vary from 0.61 to 0.82 for individuals with PD and generalised anxiety disorder, respectively (Danker-Hopfe et al. 2004). Our automated classifier achieves a similar benchmark, with a Cohen's Kappa for HCs of 0.73, and for the combined cohort of healthy and RBD participants of 0.62.

Compared to the HCs the drop in sleep staging performance in the RBD cohort is due to a greater rate of misclassification, especially with regards to N2 and REM. Annotated N2 epochs in RBD participants are misclassified for W, N3, and REM by 8.68%, 7.43%, and 2.88%, respectively (compared to 0.988%, 3.98%, and 1.48% in HCs). This misclassification for N2 is due to greater variation in EEG characteristics compared to healthy controls. This coupled with the inability to exploit diminished levels of atonia contributes to a decrease in automated sleep staging performance. The increased misclassification of annotated W epochs in the RBD cohort, compared to HCs, can partly be attributed to the greater prevalence of annotated W epochs. This is because individuals with RBD tend to have more interrupted and erratic sleep patterns. Furthermore the EMG signal helps distinguish REM from W in HCs, but this attribute is often not helpful in the context of RBD, where there can be an absence of atonia in REM. Critical to RBD diagnosis is the identification of REM sleep, and while other studies in automated sleep staging produce better results in REM sleep detection, they benefit from primarily focusing on young HCs or a relatively smaller sample size (Virkkala et al. 2008; Güneş et al. 2010; Fraiwan et al. 2012; Kempfner et al. 2012, 2013b; Liang et al. 2012; Bajaj and Pachori 2013; Khalighi et al. 2013; Imtiaz and Rodriguez-Villegas 2014; McCarty et al. 2014; Sousa et al. 2015; Lajnef et al. 2015; Yetton et al. 2016). Despite the lack of sensitivity, REM specificity remains high, which means that as long as REM is identified with a certain precision, the quantified absence of atonia will remain indicative.

The high correlation coefficients depicted in Figure 2 illustrate the success of automated sleep staging in preserving EMG metrics. From the established metrics, the motor activity metric provides the lowest correlation, which can be attributed to its dependence on several adjustable parameters such as the motor activity detection threshold, minimum motor activity event duration and the inter-event interval. Once more this technique is susceptible to sleep stage misclassification because it relies on calculating variance during REM, which is especially variable in individuals with RBD. The STREAM algorithm proves more robust and immune to high variability, because it compares variance between REM and NREM. The atonia index (REM) not only proves robust but is also parameter free (excluding window lengths) and provides a correlation scores of 0.90, one of the best amongst established metrics. Additional features that incorporate sleep architecture also remain unaffected using automated sleep staging. The atonia index ratios (N2/REM and N3/REM) have correlation scores (0.80 and 0.79, respectively) less than the traditional REM atonia index but they remain relatively high due to the sleep stage classification performance of N2 and N3. The success of the multi-stage classification also contributes to the high correlation scores for N3 sleep ratio and sleep efficiency (0.75 and 0.79, respectively). The EMG fractal exponent ratios (N2/REM and N3/REM) prove to provide the best correlation scores, with 0.99 and 0.97, respectively. Aided by the automatic multi-sleep stage classification performance, these metrics also benefit from the robustness of information derived from the frequency spectrum (specifically during REM, N2 and N3).

The performance of established metrics in RBD detection is summarised in Table 5, where results are comparable to those published in literature (Burns et al. 2007; Ferri et al. 2010; Frandsen et al. 2015). These techniques succeed because they quantify levels of both tonic and phasic EMG activity during REM, emulating the visual diagnostic inspection of clinicians. From the established metrics, the atonia index provides the best performance in RBD detection when using either manually annotated or automatically classified sleep stages. Misclassification can occur because individuals with RBD can have a normal night's sleep (free of RBD characteristics) and consequently produce a high atonia index score, mimicking that of a HC participant. This is to be expected from a disorder that displays symptoms intermittently and provides an argument towards developing a take-home PSG device that could be applied over multiple nights (Sterr et al. 2018). Conversely, there is one instance of a healthy control that produces a low atonia index due to movements captured during REM, coupled with various artefacts within the signal.

The atonia index is outperformed when all established metrics are used in combination with a RF classifier, as detailed in Table 5. While the sensitivity and accuracy of the RF classifier, trained by established metrics, outperforms individual metrics, the atonia index and STREAM metrics retain a higher specificity. The RF classifier is able to exploit the information provided by the atonia index and incorporate the added detail within STREAM and motor activity to boost sensitivity. Supplementing these established metrics with additional features (that describe the atonia index ratios, EMG fractal exponent and sleep architecture) in a RF classifier provided the best performance in RBD detection. These additional features incorporate the transitional changes in atonia between sleep stages (Kryger et al. 2011) and the altered sleep architecture that are inherently different between HCs and those with RBD (Massicotte-marquez et al. 2005). This is further substantiated by Figure 3, where the ratio of N3 sleep proves to be the second most important feature. We can also observe that the atonia index and fractal exponent ratios prove more important than motor activity and STREAM. While the ratio of N3 sleep is able to help distinguish between HCs and those with RBD, it may prove too general when applied to individuals with various other neurological and confounding sleep disorders.

Finally the effect of automated sleep staging on RBD detection is also summarised in Table 5, where the success of accurate sleep staging ensures the performances remains high. Metrics used for RBD detection are largely derived from REM sleep, the impact of misclassifying actual REM for other sleep stages means less epochs are available to analyse and detect RSWA (bias towards negative RBD detection). Inversely, other sleep stages misclassified as REM, might have the impact of biasing positive RBD detection, however due to high specificity of REM classification, this effect is nullified. Subsequently, RBD detection performance is only slightly diminished when using automatically classified sleep stages (see Table 5).

These results could be further improved upon with a better automated sleep stage classification, which might be realised with deep learning techniques (Phan et al.; Andreotti et al. 2018). Although perfect automated sleep staging may never be attainable, given human inter/intra variability, this study has demonstrated that this isn't necessarily required for RBD detection. Given sufficiently precise automated sleep staging, enough information is retained to identify RBD using an RF classifier and a combination of metrics that measure the absence of atonia, the atonia changes between different sleep stages, and sleep architecture.

## 6. Limitations & Future Direction

Although this study included a reasonably large number of RBD PSG recordings, further work is required in order to define the potential role of automated RBD detection in the clinical environment. This study is limited in focusing on distinguishing RBD individuals from HCs, and there maybe additional confounds in clinical practise when applied to a mixed population of sleep disorders. To help further delineate RBD, a promising signature might be found with the inclusion of other non-motor features, such as altered heart rate variability during sleep (Sorensen et al. 2013). It is also possible that the RBD-specific features identified here may prove useful in identifying any differences between synucleinopathy-associated RSWA from other confounding conditions such as medication-associated RSWA. Furthermore there is evidence to suggest EMG signals provided from the flexor digitorum superficials and exterior digitorum brevis best captured phasic EMG activity and might help increases the sensitivity of RSWA detection , especially in the case of mild RBD episodes (Frauscher et al. 2008). This study is also limited in that sex ratios were not matched in the study cohorts and evidence suggests there are physiological differences in sleep between sexes. Future directions of this work would look to apply improved automated sleep staging algorithms and diagnosis across a range of common sleep disorders, whilst also incorporating other PSG and physiological signals to improve RBD detection.

## 7. Conclusion

This study proposes a fully-automated pipeline for RBD detection based on a combination of established EMG metrics (atonia index, motor activity, and STREAM) that also incorporated sleep architecture. The algorithm outperforms individual metrics and demonstrates that the atonia levels between sleep stages can help distinguish RBD individuals from HCs. This study also achieved automated sleep staging with a level of accuracy comparable to manual annotation, in a large cohort of 53 aged-matched RBD and 53 HCs using a limited PSG montage.

# Table & Figures

| Database | Cohort | Age | #Subjects | #Female | #Male |
|---|---|---|---|---|---|
| CAP | RBD | 70.7±6.2 | 22 | 3 | 19 |
| JR | RBD | 63.7±7.6 | 31 | 2 | 29 |
| **Combined (CAP/JR)** | **RBD** | **66.6±7.8** | **53** | **5** | **48** |
| **MASS** | **Elderly-HC** | **63.0±5.0** | **53** | **19** | **34** |

*Table 1: Datasets used in the study. Recordings from individuals with RBD are collected from the JR and CAP datasets and have been combined (CAP/JR). Note the male predominance within the RBD dataset. These datasets are analysed separately for sleep stage classification and combined for abnormal EMG calculation and RBD detection.*

| Channel | Category | Name | Description | Reference |
|---|---|---|---|---|
| EEG | Time | Zero Crossing Rate | The number of instances where the EEG signal crosses the reference line, which itself is calculated as the mean value. | (Susmáková and Krakovská 2008) |
| EEG | Time | Hjorth Parameters | The stationary Hjorth parameters were calculated for each mini-epoch to derive features indicative of activity, mobility and complexity. These parameters are based on variance of the derivatives of the waveform. | (Motamedi-Fakhr et al. 2014) |
| EEG | Time | Time Domain Properties | The time domain properties for each mini-epoch are derived using the low-power and log amplitude of each derivative up to the 10th derivative | |
| EEG | Time | Percent Differential | This feature is calculated by taking the difference between the 75th and the 25th percentile. The 75th and 25th percentile is defined by the amplitude, below which represents 75% and 25% of the random value, respectively. | (Lajnef et al. 2015) |
| EEG | Time | Mean, Minimum, & Maximum Coastline | Calculated by summating the absolute derivatives of the signal. The coastline is determined for each mini-epoch, where the mean, minimum and maximum can be calculated for a given 30s epoch. | (Yetton et al. 2016) |
| EEG | Frequency | Short Time Fourier Transform (STFT) Magnitude | The STFT uses a windowing function of fixed width that separates the non-stationary signal into segments (considered close to stationary). The Fourier transform is then applied to the windowed signal. The frequency magnitude was calculated for each clinically relevant band. | |
| EEG | Frequency | Relative Spectral Power ($RSP_\alpha$, $RSP_\beta$, $RSP_\gamma$ and $RSP_\delta$): | Relative power ratios were calculated for the α, β, γ and δ frequency bands. The ratio was calculated by taking the ratio of the average power between each specific range and the total power. Where the total power was the summation of the individual bands ($RSP_{total} = RSP_\delta + RSP_\theta + RSP_\alpha + RSP_\beta + RSP_\gamma$). | (Susmáková and Krakovská 2008) |
| EEG | Frequency | Spectral Edge Frequency | The spectral edge frequency (SEF) is defined as the frequency below which 95% of the signal power is contained. THE SEF at 95% are usually highest during REM stages. | (Susmáková and Krakovská 2008) |
| EEG | Non-linear | Teager-Kaiser Energy Operator (δ, lower α and α) | For the δ, α and lower α (8-10Hz) frequency ranges the mean and standard deviation of the Teager-Kaiser Energy Operator (TKEO) was calculated. | (Lajnef et al. 2015) |
| EEG | Non-linear | Square-Energy Operator (SEO) | For every clinically relevant frequency band the mean and standard deviation of the square energy operator was calculated. | (Tsanas et al. 2010) |
| EOG | Time | Autocorrelation | The peak of the autocorrelation sequence of every mini-epoch was calculated. The autocorrelation gives an indication of a pattern within a given mini-epoch, enhancing REM spikes within the EOG signal. | (Yetton et al. 2016) |
| EOG | Time | Variance | The variance of each mini-epoch. | |
| EOG | Time | Max Peak | The maximum absolute peak of each mini-epoch was detected as a feature (positive or negative). | (Yetton et al. 2016) |
| EOG | Time | Second Max Peak | The second largest maximum absolute peak of each mini-epoch was detected as a feature (positive or negative). | (Yetton et al. 2016) |
| EOG | Time | Differential & Maximum Differential | The average differential value is calculated for every mini-epoch, where the average and the maximum value across the 30s epoch is determined. | (this work) |

| Channel | Category | Name | Description | Reference |
|---|---|---|---|---|
| **EOG** | Frequency | Power Ratio (0.5Hz/4Hz) | The power ratio between the 0-4Hz band and the entire spectrum was calculated. | (Susmáková and Krakovská 2008) |
| **EOG** | Wavelet | Discrete Wavelet Transform (DWT) - Haar/DB2 | The DWT was applied using 4 levels of decomposition and used the maximum amplitude of the -4 level inverse DWT as the feature. The larger the amplitude of the inverse DWT signal the more the signal can be composed of wavelets and also more likely that the window contains REM | (Yetton et al. 2016) |
| **EOG** | Non-linear | Permutation Entropy | Permutation entropy is a nonlinear measure that characterises the complexity of a time series. For every mini-epoch the permutation entropy was calculated to the 10th order. | (Lajnef et al. 2015) |
| **EEG & EOG** | Time | Cross Correlation P-Value | The Pearson cross correlation coefficients are calculated and the p-values for testing the hypothesis that there is no relationship between these two signals. | (this work) |
| **EEG & EOG** | Frequency | Magnitude-Squared Coherence | The coherence measures the level of synchrony between two signals. | (Susmáková and Krakovská 2008) |
| **EMG** | Time | Atonia Index | The distribution of amplitude values (averaged and corrected) for a given 30s epoch are used to determine the percentage of values ≤ 1µV (excluding values > 1µV and < 2µV). | (Ferri et al. 2010) |
| **EMG** | Time | Energy | For every epoch the mean rectified amplitude is calculated. | (Hsu et al. 2013) |
| **EMG** | Time | 75th Percentile | The 75th percentile, detailing the value which 75% of the variable is below. | (Charbonnier et al. 2011) |
| **EMG** | Time | Entropy | The variability of the signal is calculated as follows, where $P_i$ is the histogram count of values using 256 bins: | (Charbonnier et al. 2011) |
| **EMG** | Time | Motor Activity | An algorithm that derives a signal to measure motor activity and a threshold (baseline) to determine when there is movement. The threshold is then used to determine the duration of motor activity within each epoch. | (Frandsen et al. 2015) |
| **EMG** | Frequency | Fractal Exponent | The negative slope of the spectral density using a logarithmic on both the frequency and power. $P(f) \sim f^{-\alpha}$ | (Susmáková and Krakovská 2008) |
| **EMG** | Frequency | Absolute Gamma Power | The average power in the gamma frequency range (30-100Hz). | (Susmáková and Krakovská 2008) |
| **EMG** | Frequency | Relative Power | The ratio of the average frequency power between the high frequency range (12.5-21Hz) and the total frequency band (8-32Hz). | (Charbonnier et al. 2011) |
| **EMG** | Frequency | Spectral Edge Frequency | The highest frequency at which 95% of the total signal power is located. | (Charbonnier et al. 2011) |
| **NA** | Time | Hours Recorded | The nature of sleep architecture means the progress of time provides details on the likelihood of having REM, therefore the number of hours into PSG recordings is also used as a feature. | (this work) |

*Table 2: Features extracted from EEG, EOG, and EMG signals.*

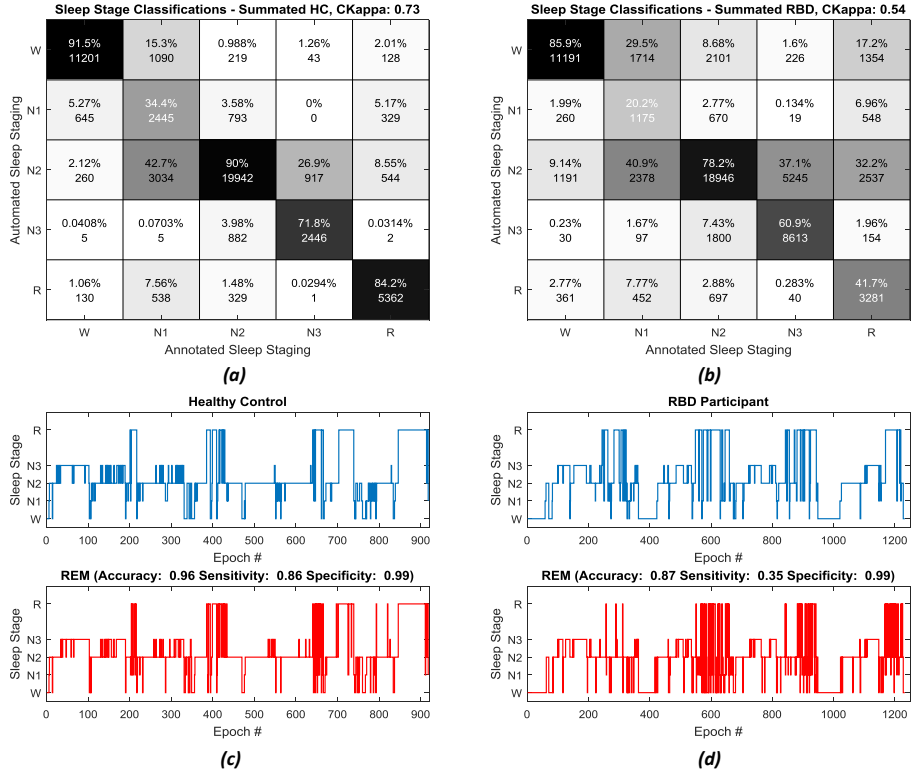

*Figure 1: (a) The confusion matrix for RBD individuals and (b) HCs, separately. The multi-stage agreement score (Cohen's Kappa) of the combined dataset was 0.62 and 0.73 and 0.54 for the individual HCs and RBD cohorts, respectively. Because individuals with RBD present REM sleep with elevated muscle tone, their REM epochs are often misclassified as Wake, unlike in HCs. An example of automated sleep staging (red) compared to manually annotated sleep staging (blue) is shown (c) and (d) for an RBD and HC participant, respectively.*

| Sleep Stage | W | N1 | N2 | N3 | REM |
|---|---|---|---|---|---|
| **Accuracy** | 0.95±0.03 | 0.88±0.05 | 0.86±0.05 | 0.96±0.03 | 0.96±0.02 |
| **Sensitivity** | 0.90±0.08 | 0.32±0.15 | 0.90±0.07 | 0.57±0.31 | 0.83±0.18 |
| **Specificity** | 0.96±0.03 | 0.96±0.04 | 0.83±0.08 | 0.98±0.02 | 0.98±0.02 |
| **Precision** | 0.87±0.10 | 0.60±0.18 | 0.80±0.08 | 0.60±0.33 | 0.84±0.15 |
| **F1** | 0.88±0.07 | 0.39±0.15 | 0.85±0.05 | 0.53±0.29 | 0.81±0.16 |

*Table 3: Performance of automatic sleep stage classification using the combined dataset but isolated for HCs.*

| Sleep Stage | W | N1 | N2 | N3 | REM |
|---|---|---|---|---|---|
| **Accuracy** | 0.89±0.09 | 0.90±0.07 | 0.75±0.10 | 0.89±0.08 | 0.91±0.05 |
| **Sensitivity** | 0.83±0.19 | 0.17±0.21 | 0.79±0.15 | 0.64±0.28 | 0.45±0.30 |
| **Specificity** | 0.89±0.12 | 0.97±0.03 | 0.73±0.14 | 0.96±0.07 | 0.97±0.04 |
| **Precision** | 0.71±0.20 | 0.36±0.29 | 0.63±0.17 | 0.84±0.18 | 0.71±0.21 |
| **F1** | 0.74±0.18 | 0.20±0.21 | 0.69±0.14 | 0.67±0.24 | 0.49±0.27 |

*Table 4: The performance of automatic sleep stage classification in RBD is substantially lower than HCs. Note in particular the reduced REM sensitivity compared to the HCs.*

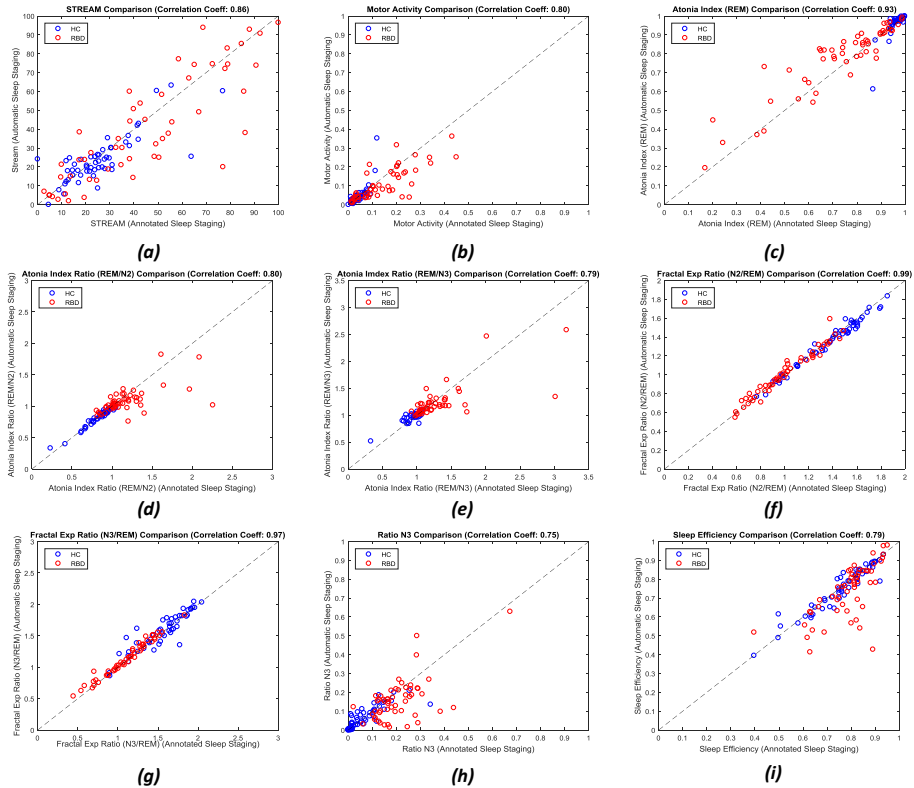

*Figure 2: A comparison of established REM EMG quantification metrics using annotated and automatically classified REM for (a) STREAM, (b) motor activity detection and (c) the atonia index. Additional features that compare levels of atonia between sleep stages and sleep architecture using annotated and automatically classified sleep stages for (d) the atonia index ratio between REM and N2, (e) the atonia index ratio between REM and N3, (f) the fractal exponent ratio between REM and N2, (g) the fractal exponent ratio between REM and N3, (h) ratio of N3 sleep, and (i) sleep efficiency. The atonia index and STREAM metrics have the highest correlation amongst established techniques but is exceeded by fractal exponent ratios.*

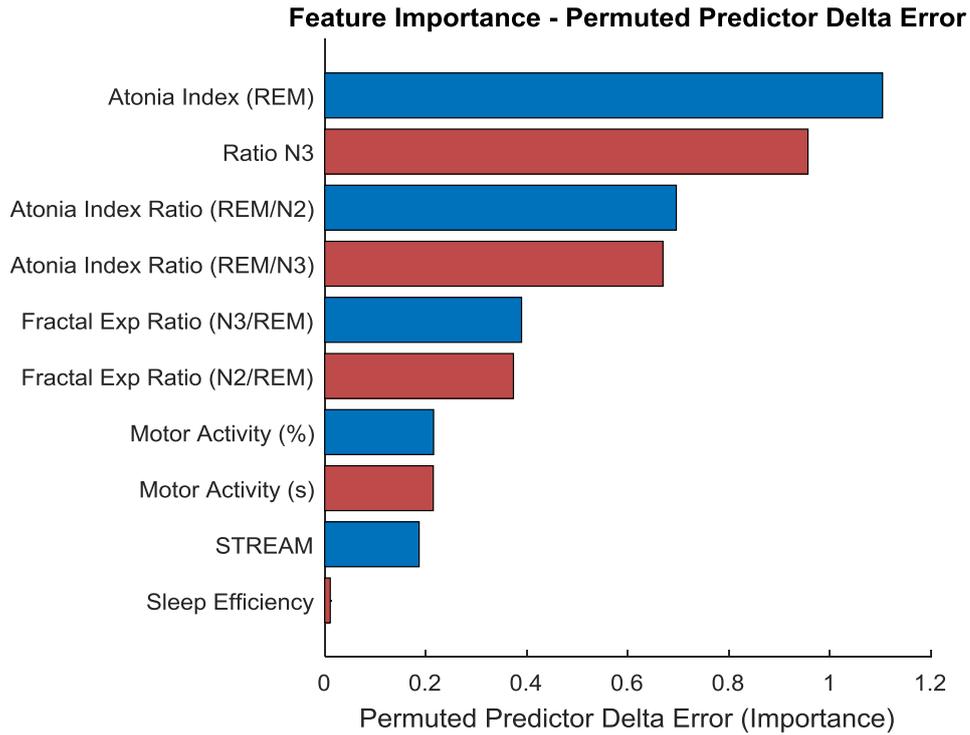

*Figure 3: Feature importance ranked using the permuted prediction error derived from the each trained random forest classifier. The atonia index during REM proved the most important, followed by the atonia index ratio (REM/N3). The ratio N3 sleep, atonia index ratio (REM/N2), atonia index ratio (REM/N3), and the fractal exponent ratio (N3/REM) also prove to be more important that established metrics such as the motor activity detection and STREAM.*

| RBD Detection | Annotated/Automatic Staging | | |
|---|---|---|---|
| | Accuracy | Sensitivity | Specificity |
| **Motor Activity** | 0.75/0.66 | 0.53/0.38 | 0.96/0.94 |
| **STREAM** | 0.70/0.68 | 0.66/0.57 | 0.74/0.79 |
| **Atonia Index** | 0.86/0.82 | 0.75/0.70 | 0.96/0.94 |
| **RF (Established Features)** | 0.91/0.91 | 0.89/0.91 | 0.92/0.91 |
| **RF (Additional Features)** | 0.96/0.92 | 0.98/0.91 | 0.94/0.93 |

*Table 5: Performance of RBD detection using atonia index, motor activity, and STREAM from manually annotated and automatically classified REM. RBD detection results are also provided using the proposed RF classifier, using established features only (atonia index, motor activity, and STREAM) and established features with additional features (atonia index ratios, N3%, sleep efficiency and REM fractal exponent). RBD detection using annotated or automatically classified sleep stages are very similar, which is a direct result of successful sleep stage classification.*